\documentclass{article} 

\usepackage{microtype}
\usepackage{graphicx}
\usepackage{subfigure}
\usepackage{booktabs} 
\usepackage{amsmath}
\usepackage{hyperref}
\usepackage[accepted]{./icml2021/icml2021_climate}
\usepackage{mathtools}

\DeclarePairedDelimiter\abs{\lvert}{\rvert}
\DeclareMathOperator*{\argmax}{argmax}

\begin{document}

    \twocolumn[
    \icmltitle{Estimation of Corporate Greenhouse Gas Emissions via Machine Learning}
    \icmlsetsymbol{equal}{*}
    \icmlsetsymbol{atbb}{*}

    \begin{icmlauthorlist}
        \icmlauthor{You Han}{bb}
        \icmlauthor{Achintya Gopal}{bb}
        \icmlauthor{Liwen Ouyang}{bb}
        \icmlauthor{Aaron Key}{aws,atbb}
    \end{icmlauthorlist}

    \icmlaffiliation{bb}{Bloomberg Quant Research, New York, NY, USA}
    \icmlaffiliation{aws}{Amazon Web Services, New York, NY, USA}

    \icmlcorrespondingauthor{You Han}{yhan137@bloomberg.net}
    \icmlkeywords{Corporate Greenhouse Gas Emissions, Climate Change, Machine Learning, ICML}
    \vskip 0.3in
    ]

    \newcommand{\workAtBloomberg}{\textsuperscript{*}Work done while at Bloomberg.}
    \printAffiliationsAndNotice{\workAtBloomberg}

    \begin{abstract}
        As an important step to fulfill the Paris Agreement and achieve net-zero emissions by 2050, the European Commission adopted the most ambitious package of climate impact measures in April 2021 to improve the flow of capital towards sustainable activities. For these and other international measures to be successful, reliable data is key. The ability to see the carbon footprint of companies around the world will be critical for investors to comply with the measures. However, with only a small portion of companies volunteering to disclose their greenhouse gas (GHG) emissions, it is nearly impossible for investors to align their investment strategies with the measures. By training a machine learning model on disclosed GHG emissions, we are able to estimate the emissions of other companies globally who do not disclose their emissions. In this paper, we show that our model provides accurate estimates of corporate GHG emissions to investors such that they are able to align their investments with the regulatory measures and achieve net-zero goals.
    \end{abstract}

    \section{Introduction}
    \label{introduction}

    There is clear, scientific evidence to show the devastating impact that increased greenhouse gas (GHG) emissions is having across the world \cite{nasa_evidence,strona2018co}. Governments and private society now recognize the urgency with which they need to act to mitigate the risks that climate change poses. Private companies have joined international governments in their pledge to hit net zero by 2050 \cite{pledge}. In order to make a meaningful change, we need to measure who is contributing GHG into the atmosphere and monitor their claims to decarbonize.

    In April 2021, the European Commission adopted the most ambitious and comprehensive package of measures to improve the investment towards sustainable finance \cite{eu_package}. For these and other international measures to be successful, it is crucial that global investors are aware of the GHG emissions of companies operating around the world. However, since GHG emissions are currently disclosed on a voluntary basis \cite{voluntary_reporting}, merely 2.27\% of companies filing financial statements are disclosing their GHG emissions according to our Environmental, Social and Governance (ESG) datasets. Moreover, the European Parliament requires investors to apply ``the precautionary principle" that penalizes non-disclosing companies by appropriately overestimating their emissions \cite{ep_precautionary}.

    There are four major challenges to estimate GHG emissions:
    \begin{itemize}
        \item \textit{Significantly imbalanced labeled and unlabeled datasets}. Here, labeled dataset means the GHG emissions disclosed by companies. Specifically, our labeled dataset contains only 24,052 annual GHG emissions disclosed by 3,960 companies from fiscal years 2000 to 2020, while our unlabeled dataset (target prediction universe) contains 619,703 records of 61,467 companies in the same fiscal range as the labeled dataset.
        \item \textit{High feature missing rates}. Although our dataset includes more than 1,000 features on public and private companies, 525 of the features have missing rates greater than 90\%.
        \item \textit{Mismatched feature missing patterns between the labeled and unlabeled datasets}. It is observed that companies who disclose their GHG emissions tend to disclose other ESG features as well. For example, energy consumption is disclosed by 97\% companies in the labeled dataset while by only 20\% companies in the unlabeled dataset.
        \item \textit{Uncertainty of estimation}. According to the precautionary principle, the uncertainty around an estimate must be quantified, which means models outputting a single point estimate of the GHG emission do not suffice.
    \end{itemize}

    The problem of estimating corporate GHG emissions has been studied in a few prior works, and their methodologies can be classified into two categories: conventional statistical approaches and modern machine learning approaches. Conventional statistical approaches either assign the mean or median of disclosed GHG emissions of an industry sector to the non-disclosing companies \cite{first_ml}, or build a simple linear regression model for each industry sector using only a few universally available features such as annual revenue of a company \cite{multivariate_regression, cdp_model, first_ml}. Such models highly rely on various assumptions that may not hold in reality such as the features are always available, all companies in the same industry sector have the same emission level, and linear relationship between the features and the GHG emissions. As an example, the Carbon Disclosure Project (CDP) applies a Gamma Generalized Linear Model (GLM) trained with 2,000 observations to estimate the GHG emissions of 5,000 non-disclosing companies \cite{cdp_model}, in which only revenues of a company from various CDP defined activities are used. Here, the GHG emissions of a company is assumed to follow a Gamma distribution without empirical justification.

    Similar to our work, two recent works apply modern machine learning approaches to model corporate GHG emissions. A Radial Basis Function network is employed in \citet{rbf_network} to estimate the GHG emissions of companies in the automotive industry using five industry specific features. In \citet{first_ml}, 13,435 observations from 2,289 global companies are used to train six base models: ordinary least square regression, elastic network, neural network, K-nearest neighbour, random forest, and extreme gradient boosting. Then the estimates of the six base models are ensembled to generate the final estimate. The main limitation of this work is that only a single value is estimated rather than a distribution making the precautionary principle difficult to apply.

    In this paper, we propose a framework of modern machine learning approaches to estimate corporate GHG emissions, which tackles all four of the aforementioned challenges. More specifically, we are able to handle missing features since decision trees are used in our model. In addition, we develop a technique called patterned dropout to solve the missing pattern mismatch issue between the labeled and unlabeled data. To account for the uncertainty quantification, we output the distribution of a company's GHG emissions in each fiscal year as our estimate, the mean of which can be used as the most accurate estimate while large quantiles can be used to fulfill the precautionary principle requirement. Lastly, different from prior works, our model is more data driven with fewer parametric assumptions made.

    \section{Datasets}\label{input_data}

    More than 1,000 features are used in our model that come from multiple datasets including ESG, industry classification, revenue segmentation by industry sectors, company financials (fundamentals), and corporate locations.

    The ESG dataset offers over 500 metrics: 297 for Environmental, 73 for Social, and 153 for Governance. The environmental metrics include carbon emissions and resource and energy use; social metrics include human rights and diversity and inclusion; governance metrics include criteria based on management structure, executive compensation, and employee relations.

    The industry classification dataset contains a hierarchical classification of industries where all industry sectors are broken down into at least four levels, with some going as deep as eight levels. The revenue segmentation dataset shows the percentage of a company's revenue from each industry sector. The fundamentals data includes information from three key accounting statements: balance sheet, income statement, and statement of cash flows.

    \section{Methodology}
    \label{methodology}

    Similar to \citet{cdp_model}, we model the Scope 1 and Scope 2 of corporate GHG emissions separately as we found empirically they are conditionally independent given the data. Scope 1 emissions are direct GHG emissions that occur from sources that are controlled or owned by an organization (e.g., emissions associated with fuel combustion in boilers, furnaces, vehicles) while Scope 2 emissions are indirect GHG emissions associated with the purchase of electricity, steam, heat, or cooling \cite{cdp_model}. Our machine learning model uses Gradient Boosting Decision Trees for Amortized Inference, Recalibration using Normalizing Flows, and Patterned Dropout for regularization. More specifically, we employ a two-phase model: first, a decision tree is used to map from features to Gamma distributions, and then a normalizing flow is used to transform the Gamma distributions to a more flexible class of distributions.

    \subsection{Amortized Inference with Gradient Boosting Trees}

    Amortized inference (or conditional probability modeling) \cite{gershman2014amortized} is used to learn a function that maps the features in a row of data to the parameters of a distribution. In this way, the function can learn a different estimate of GHG emissions per company per fiscal year. There are many choices for the function we can learn such as linear models, decision trees and neural networks. The choice among these depends on the requirements we have for the model, namely: non-linearity in the marginals and correlations, stability, interpretability, and missing values and categorical features handling. To address these needs, we decided to use gradient boosting decision trees (GBDTs, \cite{Friedman2001GB}) as they are not only able to capture complex non-linear relationships, but also more stable and explainable than neural networks \cite{LundbergLee2017,kumar2020problems}. More importantly, LightGBM \citep{LightGBM2017} is used in our model as it is able to handle missing values and categorical features more naturally.

    As we use more than 1,000 features and only have 24,052 rows of disclosed emissions, if we were to train our GBDT to convergence, the model would overfit to the labeled dataset. To prevent overfitting, we regularize via depth and minimum number of samples per leaf and use early stopping based on the log likelihood on a validation set.

    \subsection{Recalibration with Normalizing Flows}
    \label{sec:recalibration}

    Although we fit a Gamma distribution to GHG emissions, we know that real GHG emissions do not truly follow a Gamma distribution. To make the distribution more realistic, after training the GBDT on the training set, we recalibrate our distributions on the validation set. A common recalibration approach in the literature is isotonic regression \cite{kuleshov2018calibration}. However, this approach does not naturally fit the statistical framework we are using and is unable to predict many statistics such as the mean. We instead use a likelihood formulation of recalibration \citep{gopal2021normalizing}:

    \begin{enumerate}
        \item $ f_\theta = \argmax_{f_\theta} \sum_{i=1}^{N_{tr}} \log p_{f_\theta}(y^{tr}_i | x^{tr}_i) $
        \item $ g_\phi = \argmax_{g_\phi} \sum_{i=1}^{N_{val}} \log p_{g_\phi}(F_\theta(y^{val}_i | x^{val}_i)) $
    \end{enumerate}

    where $f_\theta(y|x)$ is the predicted Probability Density Function (PDF) function given $x$, $F_\theta(y|x)$ is the Cumulative Density Function (CDF) of $f_\theta(y|x)$, $\{x^{tr}, y^{tr}\}_{i=1}^{N_{tr}}$ is the training set and $\{x^{val}, y^{val}\}_{i=1}^{N_{val}}$ is the validation set. More specifically, $f_\theta$ is parameterized by a GBDT, and we parameterize $g_\phi$ with QuAR Flows, a flexible normalizing flow that can learn complicated shapes such as multimodality \citep{QuARFlows2020}. More details about normalizing flows can be found in Appendix \ref{appendix_flow}. Given this likelihood approach, the final recalibrated result is still a valid distribution on which we can compute any statistic of the distribution. Calibration results before and after recalibration are in Appendix \ref{appendix_calibration}.

    \subsection{Patterned Dropout}
    \label{sec:patterned_dropout}

    As discussed in Section \ref{introduction}, the feature missing patterns are quite different between labeled and unlabeled data such that a model trained solely on the labeled data does not necessarily perform equally well on the unlabeled data. In other words, the data distributions are different between the two datasets. To solve this problem, we augment the training data with a masked dataset that is created by applying the feature missing patterns of the unlabeled data to the original labeled data. We call this technique ``Patterned Dropout'' as the masked data is generated in such a way that features that go missing together in the unlabeled set will be masked together from the labeled data.

    Maximum Mean Discrepancy (MMD) \cite{Gretton12a} is used in our work to generate the masked data. More specifically, MMD is a non-parametric measure to compute the distance between the two sets of samples. Let $I_L$ and $I_U$ be the missing indicators of the labeled data $X_L$ and unlabeled data $X_U$, respectively. We build a masker model using neural networks that learns a missing indicator $\hat{I}_L$ conditional on $X_L$, applies it to $X_L$ to generate masked data $\hat{X}_L$, such that the MMD loss is minimized between the joint distributions $P(\hat{X}_L, \hat{I}_L)$ and $P(X_U, I_U)$. More details about our MMD technique can be found in Appendix \ref{appendix_mmd}.

    \section{Evaluation}
    \label{evaluation}

    Since the modeling technique is the same for both Scope 1 and 2, and the performance is similar, we focus our analysis on Scope 1 GHG emissions due to limited space. We perform ten-fold cross-validation. Similar to \citet{first_ml}, since our goal is to generalize to non-disclosing companies, we split our data by companies such that all observations of a company must be in only one of the sets: training, validation (for early stopping), and test. Moreover, we evaluate the performance of our model on both the original labeled (``\textit{Unmasked}") data and ``\textit{Masked}" data. Since the masked data is created by applying the feature missing patterns of the unlabeled data to the labeled data, it can be used to more accurately measure the model's performance on non-disclosing companies. Therefore, performance on the masked data is more important.

    We benchmark our model against two baseline models. As often used in conventional statistical approaches \cite{first_ml}, a simple baseline model (referenced as ``\textit{Simple}") is the mean of disclosed emissions at each industry sector starting from industry classification level 4, where if a sector has insufficient data (less than 50 data points), we fall back one level up. Moreover, to keep the model statistical, we also compute the variance and fit a sector level Gamma distribution by matching the moments. Same as the model in \citet{cdp_model}, our second baseline model (referenced as ``\textit{Baseline}") is a Gamma GLM model fit for each industry sector. More details about the second baseline model can be found in Appendix \ref{appendix_baseline}. Note that our model is also referenced as ``\textit{Recalib}" in the following analysis as the recalibrated distributions are our final outputs.

    \subsection{Performance of Identifying Heavy Emitters}

    As a practical use case, we compare the three models in a binary classification task: decide if a company is a high-intensity emitter. Carbon intensity (GHG emissions divided by revenue) has been widely used to measure a company's GHG performance \cite{hoffmann2008corporate}. For each level 1 industry sector (namely Industrials, Health Care, Technology, Financials, Materials, Real Estate, Utilities, Energy, Consumer Staples, Consumer Discretionary, Communications), we use the 90th percentile of the carbon intensity in the labeled data as the threshold, and a company's intensity is in Class 1 (high-intensity) if it is greater than the threshold and Class 0 (low-intensity) otherwise. To apply the precautionary principle, the 99th percentiles of the distributions are used to calculate the carbon intensity.

    Table \ref{tbl:precision_comparison} compares the Class 1 precision of the models, and shows that the ``Baseline" model has the best performance on the unmasked data while our model is superior on the masked data. As explained at the beginning of this section, performance on the masked data is more important as it is more representative of performance on the non-disclosing companies. We expect the ``Baseline" model to perform well on the unmasked data as it leverages full access to all of the features and tends to overfit to the unmasked data.

    \begin{table}
        \begin{center}
            \begin{tabular}{lccc}
                \toprule
                & {Simple} & {Baseline} & {Recalib} \\
                \midrule
                Unmasked Data  & 0.1620  & \textbf{0.3705}    & 0.2730  \\
                Masked Data   & 0.1620  & 0.1830  & \textbf{0.2648}  \\
                \bottomrule
            \end{tabular}
        \end{center}
        \caption{Precision Comparison}
        \label{tbl:precision_comparison}
    \end{table}
    \begin{table}
        \begin{center}
            \begin{tabular}{lccc}
                \toprule
                & {Simple} & {Baseline} & {Recalib} \\
                \midrule
                Unmasked Data  & 8854.89  & 9184.00    & \textbf{6657.00}  \\
                Masked Data   & 8854.89  & 9690.46  & \textbf{6744.02}  \\
                \bottomrule
            \end{tabular}
        \end{center}
        \caption{RMSE Comparison}
        \label{tbl:rmse_s1_comparison}
    \end{table}
    \begin{figure}
        \centerline{\includegraphics[width=\columnwidth]{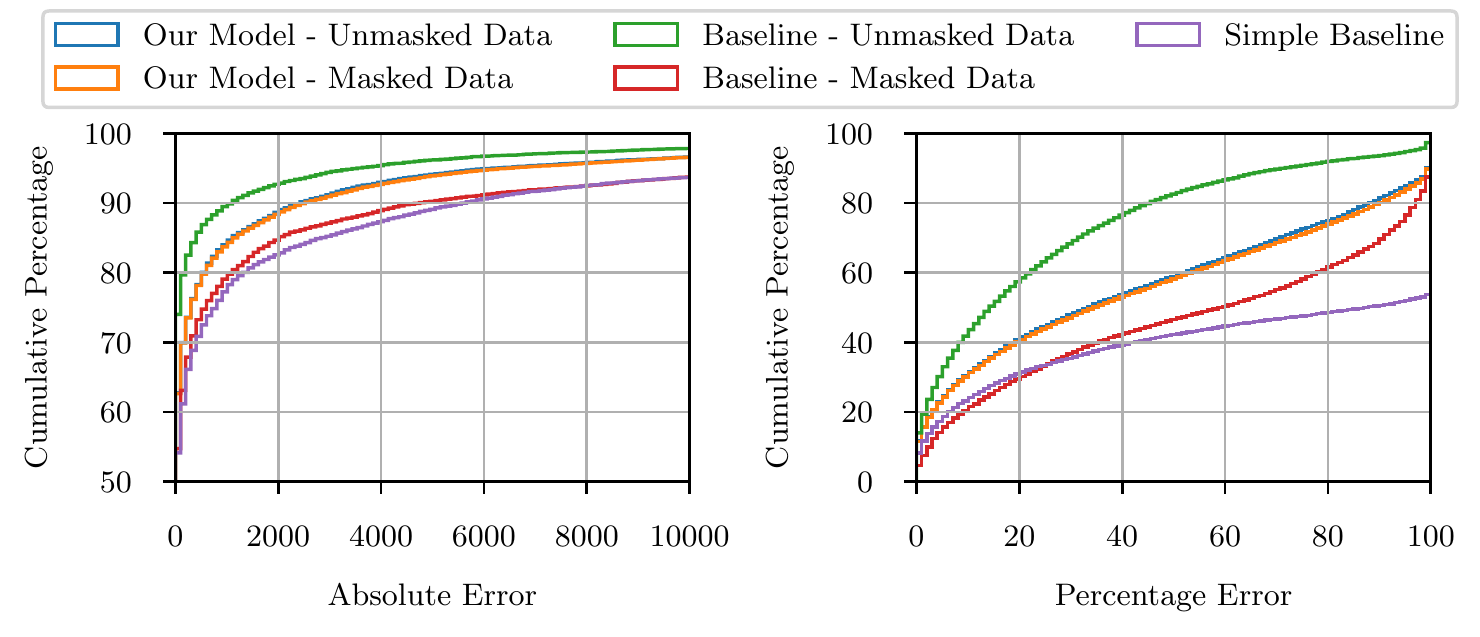}}
        \caption{Absolute and Percentage Error Comparison}
        \label{fig:error_perf_s1_sc}
    \end{figure}
    \begin{figure}
        \centerline{\includegraphics[width=\columnwidth]{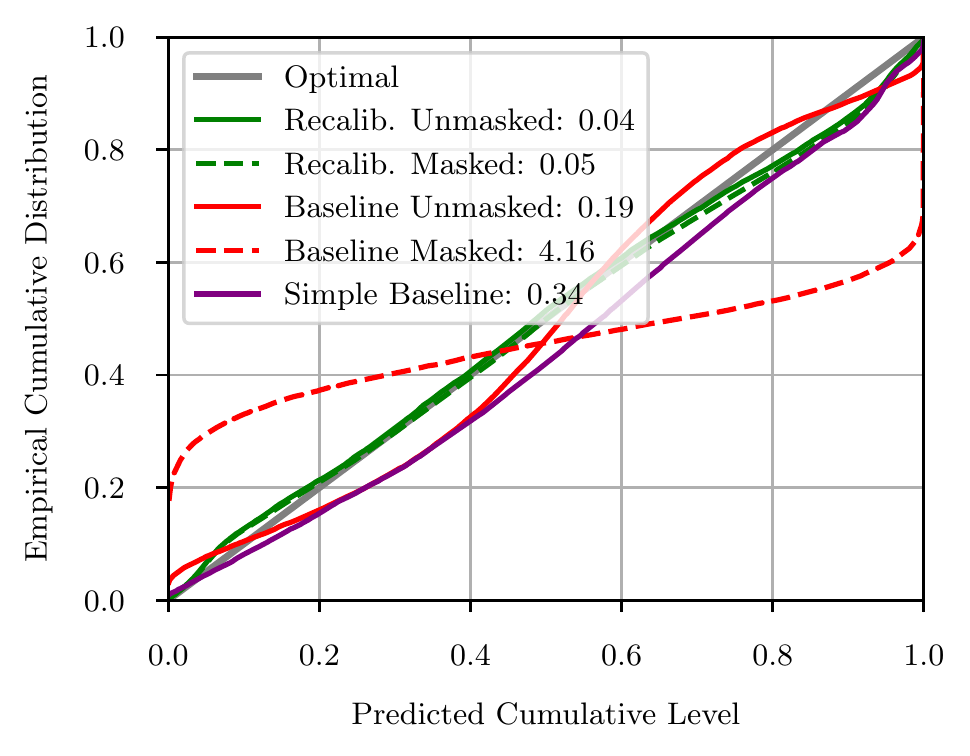}}
        \caption{CDF Q-Q Plot \cite{kuleshov2018calibration}}
        \label{fig:cdf_perf_vs_baseline}
    \end{figure}

    \subsection{Performance of Distribution Means}
    As mentioned in Section \ref{methodology}, the means of our output distributions can be used as the most accurate estimates of the GHG emissions, and they are used to evaluate the accuracy of the three models. Table \ref{tbl:rmse_s1_comparison} compares the root mean squared error (RMSE) between the baseline models and our model where our model consistently has the lowest RMSE. We also compare model performance with respect to absolute error and percentage error. As shown in Figure \ref{fig:error_perf_s1_sc}, the ``Baseline" model has lower absolute and percentage error on the unmasked data, but our model performs better on the masked data and shows less performance degradation between the unmasked and masked data.

    \subsection{Calibration Performance}

    Though a regression model is often reduced to a mean prediction, an important consideration of our model is how well-calibrated the model is. Figure \ref{fig:cdf_perf_vs_baseline} compares calibration performance of the three models, in which the calibration errors shown in the legend box are defined as $\sum_{j}(p_j - \hat{p}_j)^2$ where $p_j$ and $\hat{p}_j$ denote empirical and predicted CDFs respectively \cite{kuleshov2018calibration}.

    We can see that the ``Baseline" model is worse than our model, and the ``Simple" model is better than the ``Baseline" but worse than our model. In the baseline models, the top 10\% and the bottom 10\% of observations are not captured by the distributions, or in other words, the variance is underestimated. Moreover, our model is still well-calibrated even on masked data. The ``Baseline" model's distribution is unable to capture the top 40\% and bottom 20\% of masked observations. More results are in Appendix \ref{appendix_calibration}.

    \section{Conclusion}
    To assist investors aligning their investment strategies with EU and other international measures on Paris Agreement benchmarks, we created a machine learning model that generates distributional estimates of Scope 1 and 2 GHG emissions for non-disclosing companies. By applying the precautionary principle against those companies, regulators and investors can significantly promote sustainable finance. For future works, we plan to add more relevant features from multiple datasets such as corporate actions, supply chain and factory data.

\bibliography{./biblio}

\begin{thebibliography}{20}
\providecommand{\natexlab}[1]{#1}
\providecommand{\url}[1]{\texttt{#1}}
\expandafter\ifx\csname urlstyle\endcsname\relax
  \providecommand{\doi}[1]{doi: #1}\else
  \providecommand{\doi}{doi: \begingroup \urlstyle{rm}\Url}\fi

\bibitem[CDP(2020)]{cdp_model}
CDP.
\newblock Cdp full ghg emissions dataset technical annex iii: Statistical
  framework.
\newblock Technical report, Carbon Disclosure Project, London, UK, 2020.

\bibitem[EU({\natexlab{a}})]{ep_precautionary}
EU.
\newblock The precautionary principle: Definitions, applications and
  governance, {\natexlab{a}}.
\newblock URL
  \url{https://www.europarl.europa.eu/thinktank/en/document.html?reference=EPRS_IDA(2015)573876}.

\bibitem[EU({\natexlab{b}})]{eu_package}
EU.
\newblock Sustainable finance package, {\natexlab{b}}.
\newblock URL
  \url{https://ec.europa.eu/info/publications/210421-sustainable-finance-communication_en}.

\bibitem[Fisher-Vanden \& Thorburn(2011)Fisher-Vanden and
  Thorburn]{voluntary_reporting}
Fisher-Vanden, K. and Thorburn, K.~S.
\newblock Voluntary corporate environmental initiatives and shareholder wealth.
\newblock \emph{Journal of Environmental Economics and Management}, 62\penalty0
  (3):\penalty0 430--445, 2011.
\newblock ISSN 0095-0696.
\newblock \doi{https://doi.org/10.1016/j.jeem.2011.04.003}.
\newblock URL
  \url{https://www.sciencedirect.com/science/article/pii/S0095069611000568}.

\bibitem[Friedman(2001)]{Friedman2001GB}
Friedman, J.~H.
\newblock Greedy function approximation: A gradient boosting machine.
\newblock \emph{Ann. Statist.}, 29\penalty0 (5):\penalty0 1189--1232, 10 2001.
\newblock \doi{10.1214/aos/1013203451}.
\newblock URL \url{https://doi.org/10.1214/aos/1013203451}.

\bibitem[Gershman \& Goodman(2014)Gershman and Goodman]{gershman2014amortized}
Gershman, S. and Goodman, N.
\newblock Amortized inference in probabilistic reasoning.
\newblock In \emph{Proceedings of the annual meeting of the cognitive science
  society}, volume~36, 2014.

\bibitem[Goldhammer et~al.(2017)Goldhammer, Busse, and
  Busch]{multivariate_regression}
Goldhammer, B., Busse, C., and Busch, T.
\newblock Estimating corporate carbon footprints with externally available
  data.
\newblock \emph{Journal of Industrial Ecology}, 21\penalty0 (5):\penalty0
  1165--1179, 2017.
\newblock \doi{https://doi.org/10.1111/jiec.12522}.
\newblock URL \url{https://onlinelibrary.wiley.com/doi/abs/10.1111/jiec.12522}.

\bibitem[{Gopal}(2020)]{QuARFlows2020}
{Gopal}, A.
\newblock {Quasi-Autoregressive Residual (QuAR) Flows}.
\newblock \emph{arXiv e-prints}, art. arXiv:2009.07419, September 2020.

\bibitem[Gopal \& Key(2021)Gopal and Key]{gopal2021normalizing}
Gopal, A. and Key, A.
\newblock Normalizing flows for calibration and recalibration, 2021.
\newblock URL \url{https://openreview.net/forum?id=H8VDvtm1ij8}.

\bibitem[Gretton et~al.(2012)Gretton, Borgwardt, Rasch, Sch\"{o}lkopf, and
  Smola]{Gretton12a}
Gretton, A., Borgwardt, K.~M., Rasch, M.~J., Sch\"{o}lkopf, B., and Smola, A.
\newblock A kernel two-sample test.
\newblock \emph{Journal of Machine Learning Research}, 13:\penalty0 723--773,
  2012.

\bibitem[Hoffmann \& Busch(2008)Hoffmann and Busch]{hoffmann2008corporate}
Hoffmann, V.~H. and Busch, T.
\newblock Corporate carbon performance indicators: Carbon intensity,
  dependency, exposure, and risk.
\newblock \emph{Journal of Industrial Ecology}, 12\penalty0 (4):\penalty0
  505--520, 2008.

\bibitem[Industry()]{pledge}
Industry.
\newblock The pledge to net zero initiative.
\newblock URL \url{https://www.pledgetonetzero.org/}.

\bibitem[Javadi et~al.(2021)Javadi, Yeganeh, Abbasi, and
  Alipourmohajer]{rbf_network}
Javadi, P., Yeganeh, B., Abbasi, M., and Alipourmohajer, S.
\newblock Energy assessment and greenhouse gas predictions in the automotive
  manufacturing industry in iran.
\newblock \emph{Sustainable Production and Consumption}, 26:\penalty0 316--330,
  2021.
\newblock ISSN 2352-5509.
\newblock \doi{https://doi.org/10.1016/j.spc.2020.10.014}.
\newblock URL
  \url{https://www.sciencedirect.com/science/article/pii/S2352550920313543}.

\bibitem[Ke et~al.(2017)Ke, Meng, Finley, Wang, Chen, Ma, Ye, and
  Liu]{LightGBM2017}
Ke, G., Meng, Q., Finley, T., Wang, T., Chen, W., Ma, W., Ye, Q., and Liu,
  T.-Y.
\newblock Lightgbm: A highly efficient gradient boosting decision tree.
\newblock In Guyon, I., Luxburg, U.~V., Bengio, S., Wallach, H., Fergus, R.,
  Vishwanathan, S., and Garnett, R. (eds.), \emph{Advances in Neural
  Information Processing Systems 30}, pp.\  3146--3154. Curran Associates,
  Inc., 2017.
\newblock URL
  \url{http://papers.nips.cc/paper/6907-lightgbm-a-highly-efficient-gradient-boosting-decision-tree.pdf}.

\bibitem[Kuleshov et~al.(2018)Kuleshov, Fenner, and
  Ermon]{kuleshov2018calibration}
Kuleshov, V., Fenner, N., and Ermon, S.
\newblock Accurate uncertainties for deep learning using calibrated regression.
\newblock In Dy, J. and Krause, A. (eds.), \emph{Proceedings of the 35th
  International Conference on Machine Learning}, volume~80 of \emph{Proceedings
  of Machine Learning Research}, pp.\  2796--2804, Stockholmsmässan, Stockholm
  Sweden, 10--15 Jul 2018. PMLR.
\newblock URL \url{http://proceedings.mlr.press/v80/kuleshov18a.html}.

\bibitem[Kumar et~al.(2020)Kumar, Venkatasubramanian, Scheidegger, and
  Friedler]{kumar2020problems}
Kumar, I.~E., Venkatasubramanian, S., Scheidegger, C., and Friedler, S.
\newblock Problems with shapley-value-based explanations as feature importance
  measures, 2020.

\bibitem[Lundberg \& Lee(2017)Lundberg and Lee]{LundbergLee2017}
Lundberg, S.~M. and Lee, S.-I.
\newblock A unified approach to interpreting model predictions.
\newblock In Guyon, I., Luxburg, U.~V., Bengio, S., Wallach, H., Fergus, R.,
  Vishwanathan, S., and Garnett, R. (eds.), \emph{Advances in Neural
  Information Processing Systems 30}, pp.\  4765--4774. Curran Associates,
  Inc., 2017.
\newblock URL
  \url{http://papers.nips.cc/paper/7062-a-unified-approach-to-interpreting-model-predictions.pdf}.

\bibitem[NASA()]{nasa_evidence}
NASA.
\newblock Climate change evidence.
\newblock URL \url{https://climate.nasa.gov/evidence/}.

\bibitem[Nguyen et~al.(2021)Nguyen, Diaz-Rainey, and Kuruppuarachchi]{first_ml}
Nguyen, Q., Diaz-Rainey, I., and Kuruppuarachchi, D.
\newblock Predicting corporate carbon footprints for climate finance risk
  analyses: A machine learning approach.
\newblock \emph{Energy Economics}, 95:\penalty0 105129, 2021.
\newblock ISSN 0140-9883.
\newblock \doi{https://doi.org/10.1016/j.eneco.2021.105129}.
\newblock URL
  \url{https://www.sciencedirect.com/science/article/pii/S0140988321000347}.

\bibitem[Strona \& Bradshaw(2018)Strona and Bradshaw]{strona2018co}
Strona, G. and Bradshaw, C.~J.
\newblock Co-extinctions annihilate planetary life during extreme environmental
  change.
\newblock \emph{Scientific reports}, 8\penalty0 (1):\penalty0 1--12, 2018.

\end{thebibliography}
\bibliographystyle{./icml2021/icml2021}

    \newpage
    \appendix
    \section{Normalizing Flows}
    \label{appendix_flow}

    Learning a probability distribution from data is a core problem within machine learning. Fitting distributions can be simple for some low-dimensional datasets, but fitting distributions to high-dimensional data with complex correlations requires a more systematic solution. One deep learning approach to designing large, complex distributions that capture the essential relationships among the data points is to train Normalizing Flows. Normalizing Flows are a family of deep generative models for designing large, complex distributions that capture the essential relationships among the data points.

    Suppose that we wish to formulate a joint distribution on an $n$-dimensional real vector $x$. A flow-based approach treats $x$ as the result of a transformation $g$ applied to an underlying vector $z$ sampled from a base distribution $p_z(z)$. The generative process for flows is defined as:
    \begin{align}
        z & \sim p_z(z)
        \\ x &= g(z)
    \end{align}

    where $p_z$ is often a Normal distribution and $g$ is an invertible function. Notationally, we will use $f = g^{-1}$. Using change of variables, the log likelihood of $x$ is
    \begin{align}
        \log p_x(x) = \log p_z\left (f(x) \right) + \log \abs{\text{det}\left(\frac{\partial f(x)}{\partial x}\right)}
    \end{align}
    To train flows (i.e., maximize the log likelihood of data points), we need to be able to compute the logarithm of the absolute value of the determinant of the Jacobian of $f$, also called the \textit{log-determinant}. $\abs{\frac{\partial f(x)}{\partial x}}$ quantifies how much the function $f$ expands or contracts locally.

    Due to the required mathematical property of invertibility, multiple transformations can be composed, and the composition is guaranteed to be invertible. Since the transformations are often implemented as neural networks, the steps in the composition are easy to chain together. Thus, in theory, a potentially complex transformation can be built up from a series a smaller, simpler transformations with tractable log-determinants.

    Constructing a Normalizing Flow model in this way provides two obvious applications: drawing samples using the generative process and evaluating the probability density of the modeled distribution by computing $p_x(x)$. These require evaluating the inverse transformation $f$, the log-determinant, and the density $p_z(z)$. In practice, if inverting either $g$ or $f$ turns out to be inefficient, then one or the other of these two applications can become intractable. For the second application in particular, computing the log-determinant can be an additional trouble spot.

    \section{Maximum Mean Discrepancy}
    \label{appendix_mmd}

    Machine learning models often work on the assumption that the training and test sets follow the same distribution. In practice, this might not be the case, yet the difference in distribution, called dataset shift, is often ignored. Though this might be acceptable when the dataset shift is small, it can lead to poor performance in other cases. Dataset shift can be divided into three main categories: covariate shift, prior probability shift, and concept shift. Covariate shift occurs when the distribution of input features changes, prior probability shift occurs when the distribution of target variables changes, and concept shift occurs when the relationship between input features and target variables changes.

    Assume we have two sets of samples \(X = \{x_{i}\}_{i=1}^{N}\) and \(Y = \{y_{j}\}_{j=1}^{M}\) drawn from two distributions \(P(X)\) and \(P(Y)\). MMD is a non-parametric measure to compute the distance between the two sets of samples in mean embedding space. Let \(k\) be a measurable and bounded kernel of a reproducing kernel Hilbert space (RKHS) \(\mathcal{H}_k\) of functions, then the empirical estimation of MMD between the two distributions in \(\mathcal{H}_k\) can be written as
    \begin{align}
        \mathcal{L}_{MMD^{2}}(X,Y)\ & = \frac{1}{N^{2}}\sum_{i=1}^{N}\sum_{i'=1}^{N} k(x_{i},x_{i'}) \nonumber \\
        &+ \frac{1}{M^{2}}\sum_{j=1}^{M}\sum_{j'=1}^{M} k(y_{j},y_{j'}) \nonumber \\
        &- \frac{1}{NM}\sum_{i=1}^{N}\sum_{j=1}^{M} k(x_{i},y_{j})
    \end{align}

    When the underlying kernel is characteristic, MMD is zero if and only if \(P(X) = P(Y)\) \citep{Gretton12a}. For example, the popular Gaussian RBF kernel, \(k(x,x') = \exp(-\frac{1}{2\sigma^2} |x-x'|^{2})\), is a characteristic kernel and was widely used.

    In our MMD masker model, let \(I_{tr}\) and \(I_{te}\) be the missingness indicators for the training (i.e., labeled dataset) and test (i.e., unlabeled dataset) features \(X_{tr}\) and \(X_{te}\), respectively; we want to learn a missingness indicator \(\hat{I}_{tr}\) conditioned on \(X_{tr}\) and \(I_{tr}\) that can minimize the MMD loss between the joint distributions \(P(\hat{X}_{tr},\hat{I}_{tr})\) and \(P(X_{te}, I_{te})\):

    \begin{equation} \label{eq:mmd_mask_loss}
    \mathcal{L} = \mathcal{L}_{MMD^{2}}((\hat{X}_{tr},\hat{I}_{tr}),(X_{te}, I_{te}))
    \end{equation}
    where \(\hat{X}_{tr}\) is generated by masking original $X_{tr}$ using learned $\hat{I}_{tr}$. The downstream model is then trained using \(\hat{X}_{tr}\) instead of \(X_{tr}\). In this way, the downstream model can focus on the right features for each training sample.

    \begin{figure*}
        \centerline{\includegraphics[width=0.8\textwidth]{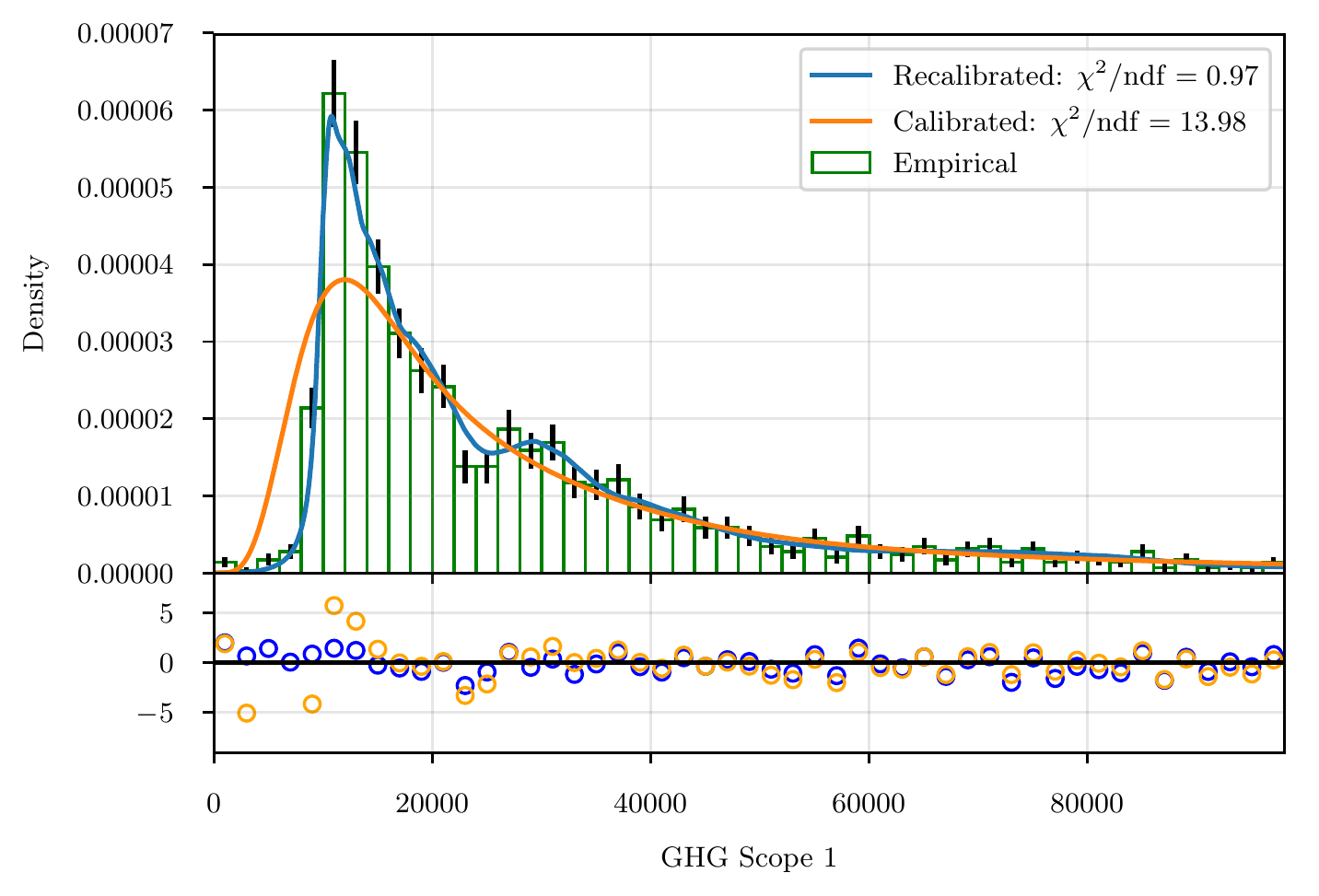}}
        \caption{Comparison of the predicted mixture distribution for before and after recalibration versus observations. The lower panel shows the distance between the predicted density and the observed density, normalized by the standard error of the counts in the bin.}
        \label{fig:empirical_vs_predicted}
    \end{figure*}

    \section{Baseline Models}
    \label{appendix_baseline}

    Our second baseline model is a linear model trained on bucketed data. The linear model we use is a Gamma Generalized Linear Model (GLM) since the distribution our GBDT fits is a Gamma distribution. We bucket the data by sectors from industry classification hierarchy, using levels from 4 to 1 (highest level). Similar to the first baseline model, if an industry has insufficient data (less than 50 data points), we fall back one level up in the industry classification hierarchy. Since we have over 1,000 features, we use a greedy approach for feature selection where we add the feature that improves the log likelihood the most. We continue the greedy selection up until the Bayesian Information Criterion increases. To add some non-linearity to our baseline model, for every positive feature, we create a new feature that is the natural log of the original feature.

    \section{Calibration Performance}
    \label{appendix_calibration}

    Figure \ref{fig:empirical_vs_predicted} compares the predicted and observed mixture distribution. Whereas the mixture before recalibration is rather smooth, the mixture from the recalibrated distribution is both qualitatively and quantitatively a better fit. The quantitative information is shown in the lower panel, where the pulls (the distance between the predicted density and the observed density, normalized by the standard error of the counts in the bin) are closer to $0$, and the $\chi^2/\text{ndf} = \sum_i \text{pull}_i^2$ is closer to one (the ideal value of the $\chi^2/\text{ndf}$). In summary, the calibration performance of our model is significantly improved through recalibration.

\begin{figure}
        \centerline{\includegraphics[width=\columnwidth]{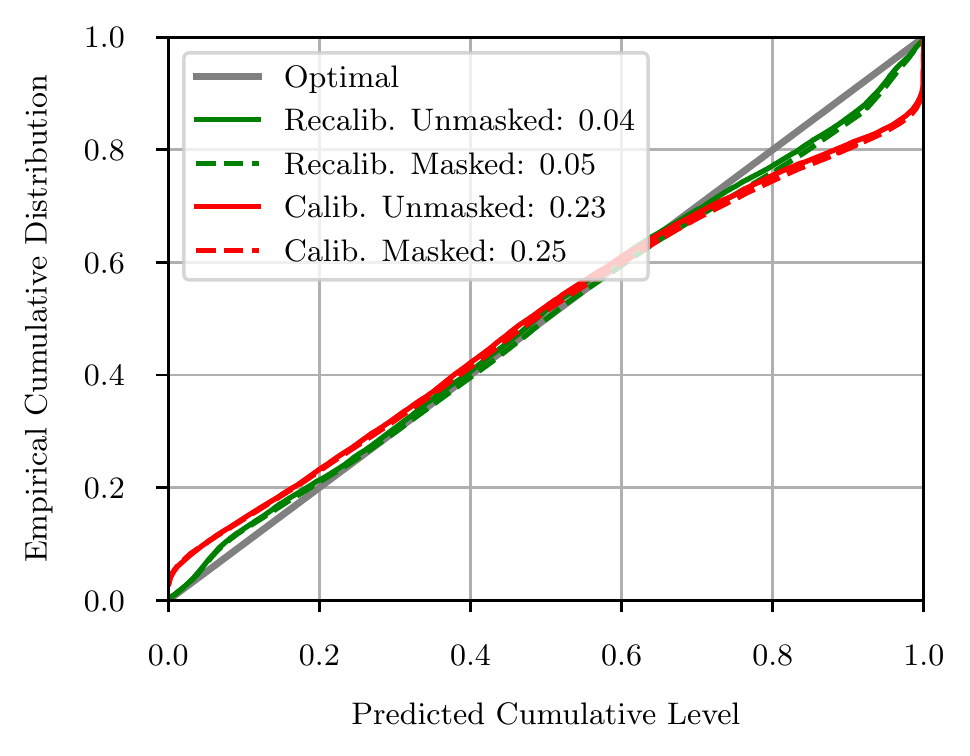}}
        \caption{CDF Q-Q Plot}
        \label{fig:cdf_recalib_vs_uncalib}
    \end{figure}

    In addition, the calibration errors are compared in Figure \ref{fig:cdf_recalib_vs_uncalib} before (referenced as ``\textit{Calib.}") and after recalibration on both the unmasked and masked data. We can see that the recalibration step decreases the calibration errors, especially on the top and bottom 10\% of observations.

\end{document}